\documentclass{article} 
\usepackage{iclr2025_conference_preprint,times}

\usepackage{amsmath,amsfonts,bm}

\def\eqref#1{Equation~\ref{#1}}

\def\1{\bm{1}}

\DeclareMathAlphabet{\mathsfit}{\encodingdefault}{\sfdefault}{m}{sl}
\SetMathAlphabet{\mathsfit}{bold}{\encodingdefault}{\sfdefault}{bx}{n}

\newcommand{\E}{\mathbb{E}}

\usepackage{graphicx} 
\usepackage{subfig}
\usepackage{caption}
\usepackage{amsmath}
\usepackage{amssymb}
\usepackage{amsthm}
\usepackage{hyperref}
\usepackage{url}
\usepackage{tikz}
\newcommand{\norm}[1]{\left\lVert#1\right\rVert}
\usepackage{hyperref}
\usepackage{url}
\newtheorem{theorem}{Theorem}[section]

\newtheorem{assumption}[theorem]{Assumption}

\title{On the Relation Between Linear Diffusion and Power Iteration}

\author{%
  Dana Weitzner$^1$, 
  Mauricio Delbracio$^2$,
  Peyman Milanfar$^2$,
  Raja Giryes$^1$\\
$^1$Tel Aviv University $^2$Google\\
\texttt{danaweitzner@mail.tau.ac.il, raja@tauex.tau.ac.il}, \\
\texttt{\{mdelbra, milanfar\}@google.com}
}

\iclrfinalcopy 
\begin{document}

\maketitle

\begin{abstract}
Recently, diffusion models have gained popularity due to their impressive generative abilities. These models learn the implicit distribution given by the training dataset, and sample new data by transforming random noise through the reverse process, which can be thought of as gradual denoising. In this work, we examine the generation process as a ``correlation machine'', where random noise is repeatedly enhanced in correlation with the implicit given distribution. 
To this end, we explore the linear case, where the optimal denoiser in the MSE sense is known to be the PCA projection. This enables us to connect the theory of diffusion models to the spiked covariance model, where the dependence of the denoiser on the noise level and the amount of training data can be expressed analytically, in the rank-1 case.
In a series of numerical experiments, we extend this result to general low rank data, and show that low frequencies emerge earlier in the generation process, where the denoising basis vectors are more aligned to the true data with a rate depending on their eigenvalues. This model allows us to show that the linear diffusion model converges in mean to the leading eigenvector of the underlying data, similarly to the prevalent power iteration method. 
Finally, we empirically demonstrate the applicability of our findings beyond the linear case, in the Jacobians of a deep, non-linear denoiser, used in general image generation tasks.
\end{abstract}

\section{Introduction}
Recently, diffusion models have gained much popularity as very successful generative models, showcasing impressive performance in image generation tasks \citep{dhariwal2021diffusion, ddpm, song2019generative, song2020score}. These models learn the implicit distribution given by the training dataset, and sample new data by transforming random noise inputs through a reverse diffusion process, which can be thought of as gradual denoising. More formally, it has been shown in \citet{kadkhodaie2023generalization} that learning the underlying distribution is equivalent to optimal denoising at all noise levels. 

In order to shed more light onto the mechanism behind the success of diffusion models, in this work we analyze the behavior of denoisers in the context of image generation, where pure noise is gradually processed into a sample from a given (implicit) distribution by gradual denoising. Unlike other works, e.g. \citet{kadkhodaie2023generalization}, we focus on the denoiser(s) throughout the generation process, and not only on the final generated data. 

To this end, we suggest the following simple model to illustrate our point. 
Consider the class of linear denoisers, where the optimal denoiser is given by a PCA projection. To simulate the diffusion generation process, we learn a series of projections onto noisy data at different noise levels, and use them to transform pure noise into samples from the underlying distribution. Given this simple model we can inspect the evolution of eigenvectors spanning gradual projections with decreasing noise levels, as well as the distribution of the generated data samples. 

We show that the correlation of the noisy basis eigenvectors with their clean version decays as the noise level increases, with a rate determined by the eigenvalues and the size of the training dataset. In other words, we show that low frequencies, corresponding to large eigenvalues, emerge earlier in the reverse process as was empirically observed in \citep{ddpm}, and analyze how more training data contributes to generalization \citep{kadkhodaie2023generalization}. Analytically, this corresponds to the spiked covariance model \citep{johnstone2001distribution}, in which we bound this decay for the leading eigenvector (corresponding to the largest eigenvalue).

Next, we demonstrate the applicability of our findings to more general, non-linear deep denoisers. Although the network is not linear, its application can be written as a linear operation of the Jacobian calculated on the input image. We empirically show that the aforementioned decay of eigenvector correlations is prevalent also in in the Jacobians of a deep denoiser, in the final stages of image generation, thus showing the relevance of our analysis in a broader context, and not just in a simplified linear case.

\section{Background and Related Work}
Since their introduction by \citet{sohl2015deep}, diffusion models have been vastly used in image generation tasks \citep{dhariwal2021diffusion, ddpm, song2019generative, song2020score}, more general computer vision tasks \citep{amit2021segdiff, baranchuk2021label, brempong2022denoising, cai2020learning}, and in other domains such as natural language processing \citep{austin2021structured, hoogeboom2021argmax, li2022diffusion, savinov2021step, yu2022latent} and temporal data modeling \citep{alcaraz2022diffusion, chen2020wavegrad, kong2020diffwave, rasul2021multivariate, tashiro2021csdi}. On top of their practical success, different flavors of training and sampling have risen based on interesting theoretical reasoning, e.g., considering the statistical properties of the intermediate data \citep{ddim, sohl2015deep}, or by framing the problem in the form of stochastic differential equations (SDEs) \citep{karras2022elucidating, song2021maximum, song2020score, chen2023geometric} or score based generative models \citep{song2019generative, song2020improved}. In this work, we look at diffusion models in the context of iterative denoising, and focus on the properties of the learned denoiser \citep{milanfar2024denoising}. 

Recently \citet{kadkhodaie2023generalization} showed that the learned denoising functions are equivalent to a shrinkage operation in a basis adapted to the underlying image. In this sense the diffusion denoiser is an adaptive filter \citep{milanfar2013atour, talebi2014nonlocal, talebi2016fast}. While they focus on the analysis of the nonlinear denoiser at the point of the final generated data, we are interested in the evolution (adaptation) of the denoiser throughout the generation process, and its dependence on the noise level. To this end, we suggest a simple linear denoising model, presented in Section \ref{sec:linear_diffusion}. In this case, the (optimal) denoiser does depend on the underlying image, and its dependence on the noise level can be traced analytically, as we show hereafter.

Due to their phenomenal empirical success, some attempts have been devoted towards providing theory supporting the sample and iteration complexity of diffusion models. The current body of work can be generally parted to attaining iteration complexity bounds assuming approximately accurate scores \citep{iteration_complexity_li2024sharp, iteration_complexity_li2023towards, iteration_complexity_chen2023restoration, iteration_complexity_huang2024convergence, iteration_complexity_benton2023linear}, and to assessing the sample complexity to learn the score functions \citep{chen2023score, block2020generative, biroli2023physics}. 
Among these works, many assume a low dimensional data distribution \citep{low_dim_debortoli2022convergence, low_dim_li2024adapting, low_dim_oko2023diffusion, chen2023score, wang2024diffusion}, which is a reasonable assumption in theoretical works. Yet, it might particularly explain the gap between the current iteration bounds and the much lower complexity apparent in practice \citep{low_dim_li2024adapting}.
In our work, we consider linear models and deduce a linear sample complexity bound associated with learning the score function in Sec. \ref{sec:basis_perturbation} and discuss the tradeoffs of the synthesis conversion rate in Sec. \ref{sec:output_distribution}. 
The previous works mentioned above mainly develop bounds assuming specific samplers and scaling details, which differ from our setting. In addition, they generally bound the Total Variation distance (under varying assumptions on the target distributions), which is not trivial to translate to the generated covariance matrix that we focus on even in the linear Gaussian case \citep{devroye2018total}. The difference in our setting enables us to connect the theory of diffusion models to a broad body of work concerning the spiked covariance model \citep{johnstone2001distribution}, and supports the analysis of denoising diffusion as a correlation machine, which is the main purpose of this paper.

In the setting of Statistical Mechanics, \citet{biroli2023physics} analyse diffusion models in very large dimensions, focusing on the Curie-Weiss model of ferromagnetism. As an introduction to their work, they also discuss a simple linear score model, in the context of the sample complexity of learning the score function. They focus their discussion on the case of Gaussian data, where the eigenvalues of the covariance matrices can be typically characterised. Unlike their work, we consider data that reside in a low dimensional subspace, with no specific distribution, described in Sec. \ref{sec:basis_perturbation}.

\section{Linear Diffusion - Problem Setup}\label{sec:linear_diffusion}
For our analysis, we define the following simple iterative linear generation model. First, define the standard diffusion model.
Let $q$ denote the natural data distribution and let $x_0 \sim q$ be a sample from the natural data ($x \in \mathbb{R}^d$). The forward (diffusion) process is defined \citep{ddpm} by
\begin{equation}
    q(x_t | x_{t - 1}) = \mathcal{N}(\sqrt{1 - \beta_t} x_{t - 1}, \beta_t \mathbf{I}) 
\end{equation}
for some fixed noise schedule $\{ \beta_t \}_{t=1}^T$ and $x_0 \sim q$. It can be shown that 
\begin{equation}
    q(x_t | x_0) = \mathcal{N}(\sqrt{\bar{\alpha}_t} x_0, (1 - \bar{\alpha}_t) \mathbf{I}),
\end{equation}
where $\alpha_t = 1 - \beta_t$ and $\bar{\alpha}_t = \Pi_{s = 1}^t \alpha_s$.
For our simplified model, consider the process (without scaling),
\begin{equation}\label{eq:diffusion_model}
    q(x_t | x_{t - 1}) = \mathcal{N}(x_{t - 1}, \sigma^2_t \mathbf{I}).
\end{equation}
This implies that $x_t = x_{t - 1} + \epsilon_{\sigma_t}$, where $\epsilon_{\sigma_t} \sim \mathcal{N}(0, \sigma^2_t \mathbf{I})$ for some fixed noise schedule $\{\sigma_t \}_{t=1}^T$. We discard the scaling to comply with previous analysis of the spiked covariance model \citep{nadler2008finite} (more details in Section \ref{sec:basis_perturbation}). This corresponds to the "Exploding Variance" formulation, used with Langevin dynamics to sample data as a variant of score based diffusion models \citep{song2019generative, song2020score, song2020improved}. We choose to present the "standard" diffusion models in the setting of denoising diffusion \cite{ddpm} and not using the score-based approach entirely, as we focus our discussion on the qualities of the denoiser.

The reverse (generation) process is defined using a parameterized distribution model $p_\theta$, generally defined by the Markov process
\begin{align}
    & p_\theta (x_{0:T}) = p(x_T) \Pi_{t = 1}^T p_\theta (x_{t - 1} | x_t), \\
    & p_\theta (x_{t - 1} | x_t) \triangleq \mathcal{N}(\mu_\theta(x_t, t), \Sigma_\theta(x_t, t)),
\end{align}
where $p(x_T) = \mathcal{N}(0, \mathbf{I})$. By choices of parametrization and loss manipulations (see \citep{ddpm}), one generally learns to estimate the error $\epsilon_\theta (x_t, t)$, where 
\begin{equation}
    \mu_\theta(x_t, t) = \frac{1}{\sqrt{\alpha_t}} \Big{(} x_t - \frac{\beta_t}{\sqrt{1 - \bar{\alpha}_t}} \epsilon_\theta (x_t, t) \Big{)},
\end{equation}
$\Sigma_\theta(x_t, t) = e^2_t \mathbf{I}$, and $e_t$ is a designed schedule (usually chosen to be equal to $\sigma_t$). Thus, the reverse process can be expressed as a denoising chain
\begin{equation}
    D_t(x_t) = \frac{1}{\sqrt{\alpha_t}} \Big{(} x_t - \frac{\beta_t}{\sqrt{1 - \bar{\alpha}_t}} \epsilon_\theta (x_t, t) \Big{)} + e_t z,
\end{equation}
where $z \sim \mathcal{N}(0, \mathbf{I})$ and $z_1 = 0$. This is a stochastic denoiser, which preserves the Markovian property of the forward process. Later versions suggested similar (non Markovian) deterministic denoisers, e.g., DDIM \citep{ddim}, or more general stochastic denoiser chains, for a continuous forward model (InDI \citep{indi}). 

In our case, we restrict the denoisers to be a linear function of $x_t$. Thus, the optimal denoiser (in the $\ell_2$ sense) is given by the PCA projection onto the target distribution. For the reverse process, we learn a simple PCA denoiser (projection) based on $X_{t - 1}$, which is the cleaner version of the training set $X = \{x_1, ..., x_n \}$ at time $t - 1$. Thus, at each time step we learn
\begin{equation}\label{eq:D_t}
    D^t_{PCA}(x_t) = D^t_{PCA}(x_t; X_{t - 1}) = P_t(x_t ; X_0 + E_{\bar{\sigma}_t}),
\end{equation}
where each column in $E_{\bar{\sigma}_t}$ is distributed by $\mathcal{N}(0, \bar{\sigma}^2_t \mathbf{I})$ and $\bar{\sigma}_t$ is a function of $\{\sigma_s \}_{s=1}^t$. Our simple denoising procedure is based on the sequential application of $P_t \in \mathbb{R}^{d \times d}$, which is the projection on perturbed principal components with respect to the clean data distribution $q$. It is a deterministic denoiser given the sampling of training data and noises, which does not depend neither on $x_t$ nor on $x_0$. Nevertheless, this model is relevant in differentiable environments of more complex settings such as DNN based denoisers, as we show in Section \ref{sec:zero_bias_denoisers}. 

\textbf{Empirical Demonstration of a Linear Diffusion Model.}
To illustrate the forward and backward processes in the linear case, we perform a numerical simulation using the MNIST dataset, which is simple enough to be estimated via a linear model. We start here with the training and generation procedures, and use the same setting and trained denoiser to demonstrate our findings throughout the paper.

\begin{figure}
    \centering
    \includegraphics[width=\textwidth]{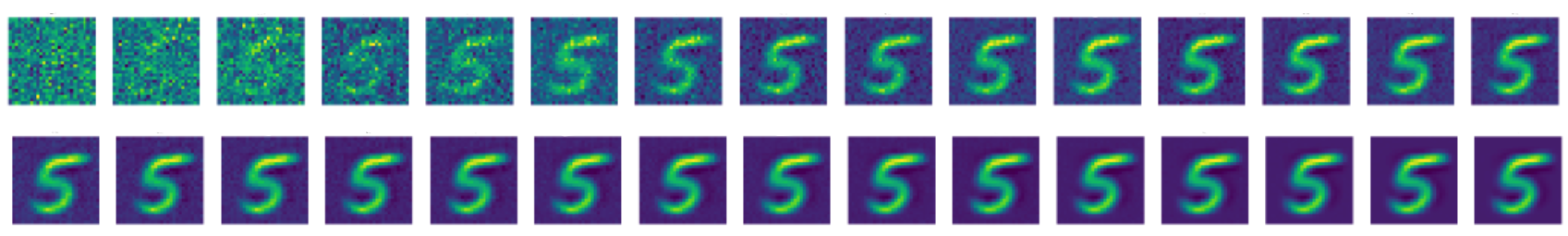}
    \caption{Digit generation from pure noise (class conditioned). The reverse process runs from left to right, top then bottom.}
    \label{fig:digit_generation}
\end{figure}

In the following experiment we simulate the process described above using the MNIST dataset (we use the default train / test splits). In the class conditioned case, we learn a PCA denoiser with 30 components for each time step where $x_t = x_{t - 1} + \epsilon_{\sigma_t}$, $\sigma_t \propto t$, and $T = 65$ iterations. Figure \ref{fig:digit_generation} shows a (decimated) example of digit generation from pure noise, where we apply the sequence of denoisers $D^t_{PCA} = P_t$, which will be more accurately defined in Section \ref{sec:basis_perturbation}. 
In order to understand the reverse process, we now turn to analyze the gradual change of $P_t$, that might be expressed by the angle between the clean and noisy components over time.

\section{Linear Diffusion as Basis Perturbation}\label{sec:basis_perturbation}
We now turn to analyze the linear model presented above and show how the generation process can be seen as a kernel ``correlation machine''. Specifically, we are interested in the temporal (i.e., noise level) dependence of the denoiser \eqref{eq:D_t} throughout the generation process.
Recall that at each time step $x_t = x_{t - 1} + \epsilon_{\sigma_t}$, where $\epsilon_{\sigma_t} \sim \mathcal{N}(0, \sigma^2_t \mathbf{I})$ (\eqref{eq:diffusion_model}). Since the noise is assumed to be Gaussian, we can write $x_t = x_0 + \epsilon_{\bar{\sigma}_t}$, where $\bar{\sigma}_t = \sqrt{\sum_{i=0}^t \sigma_i^2}$. Assume that the data distribution is such that its population covariance is given by 
\begin{align} \label{eq:spiked_model}
    \Sigma_t &= \E x_0 x_0^\dag + \bar{\sigma}^2_t \mathbf{I} 
             = \sum_{i = 0}^{r - 1} \lambda^2_i u_i u_i^\dag + \bar{\sigma}^2_t \mathbf{I} 
             \triangleq \Sigma_0 + \bar{\sigma}^2_t \mathbf{I},
\end{align}
where $r - 1 < d$, i.e., the data reside in a low dimensional subspace (which is generally true for natural data). This is known as the ``spiked model'' \citep{johnstone2001distribution}, with a vast body of work covering the distribution and identifiability of the spikes spectrum (e.g., \citep{nadler2008finite}). Throughout the paper, we use the term "index" to refer to to the index $i$ in \ref{eq:spiked_model}, where the eigenvalues $\lambda_i$ are ordered largest to smallest.

Given $n$ samples concatenated as columns in the matrix $X_0$, at each time step we learn the PCA basis associated with $X_t = X_0 + E_{\bar{\sigma}_t}$, by the diagonalization of the sample covariate matrix
\begin{align}
    \hat{\Sigma}_t &= \frac{1}{n} X_t X_t^\dag = \frac{1}{n} (X_0 X_0^\dag + X_0 E_{\bar{\sigma}_t}^\dag + E_{\bar{\sigma}_t} X_0^\dag + E_{\bar{\sigma}_t} E_{\bar{\sigma}_t}^\dag) \triangleq U_t S_t U_t^\dag.
\end{align}
Thus, during the reverse process, at each time step we apply the projection
\begin{equation}\label{eq:pca_denoiser}
    D^t_{PCA} = P_t = U_t U_t^\dag.
\end{equation}
In order to understand the diffusion generation process, we analyze the decay of the product $\langle u_i^t, u_i \rangle$ over time. Note, that there are two drivers of change in the perturbation of $u_i$ to $u_i^t$. The first being the added noise, i.e., $\norm{\Sigma_t - \Sigma_0}$. This is the key in the diffusion process and our main focus. The second, is in the finite sample approximation $\norm{\hat{\Sigma}_t - \Sigma_t}$. This source of error is interesting in the context of sample complexity, as it encompasses the approximation of the denoiser learned from a finite dataset, the equivalent of the sample complexity of learning the score function in \citep{chen2023score, block2020generative, biroli2023physics}. 

\begin{figure}
    \centering
    \includegraphics[width=0.4\textwidth]{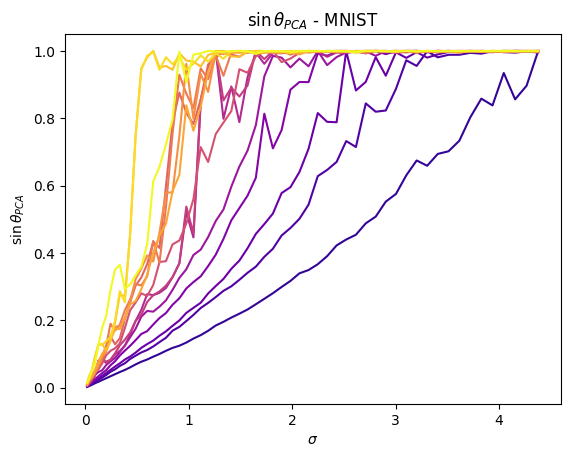}
    \caption{The sine of the angle between the clean principal components and their noisy versions, colored by the order of the eigenvalues (the darkest being largest eigenvalue). Low frequencies emerge earlier in the generation process (at higher noise levels). This motivates Assumption \ref{assumption:diagonal_elements}, that extends \eqref{eq:sin_theta_pca_0} to higher ranks.}
    \label{fig:sin_theta_pca}
\end{figure}

For the rank-1 case, \citep{nadler2008finite} presented a finite sample theorem which holds with high probability for the closeness between the leading eigenvalue and eigenvector of sample and population PCA under a spiked covariance model similar to \eqref{eq:spiked_model}. They bound the angle between the leading empirical eigenvector and its population counterpart with approximately $\mathcal{O}(d)$ sample complexity, and a linear dependence on the noise level:
\begin{equation}\label{eq:sin_theta_pca_0}
    \E \sin{\theta_\text{PCA}} = 
    \E \sqrt{1 - \langle u^t, u \rangle^2} \approx \frac{\bar{\sigma}_t}{\lambda} \sqrt{\frac{d}{n}},
\end{equation}
where $\bar{\sigma}_t$ is assumed to be small and $d \gg 1$. This result shows that the leading eigenvector rotates in a rate proportional to the noise level. Our numerical experiments on the MNIST dataset (detailed in Section \ref{sec:output_distribution}) show that this is a good approximation in practice, also for the rank-r case (Fig. \ref{fig:sin_theta_pca}).

Notice that in \eqref{eq:sin_theta_pca_0} the angle is inversely linked to the eigenvalue, inferring a slower change with higher eigenvalues. In the reverse process, we gradually move from pure noise or high noise levels to smaller noise variance. Given the lower slope of the components corresponding to larger eigenvalues, we interpret the result in Fig. \ref{fig:sin_theta_pca} as the earlier emergence of low frequencies in the generation process. The first component to be visible in the generated image is the one with the largest eigenvector, as it is the first one that shows a correlation in high noise levels. Throughout the generation process, when the noise level decreases, the next components take presence, by the order of their associated eigenvalue - from the larger to the smaller. Finally, the components with the smallest eigenvalues appear when the noise level is low. 

\begin{figure}
\centering
    \subfloat{\includegraphics[width=0.3\linewidth]{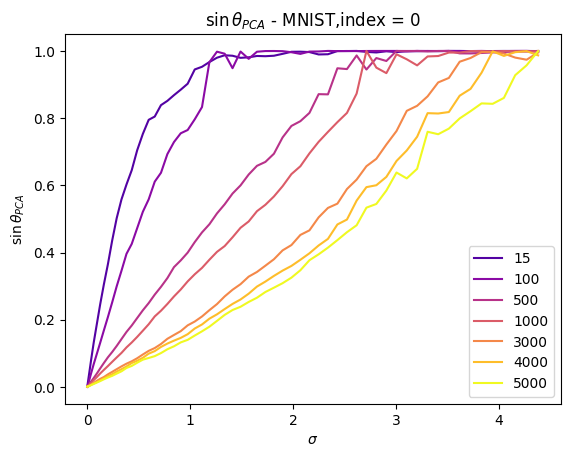}}
    \hfil
    \subfloat{\includegraphics[width=0.3\linewidth]{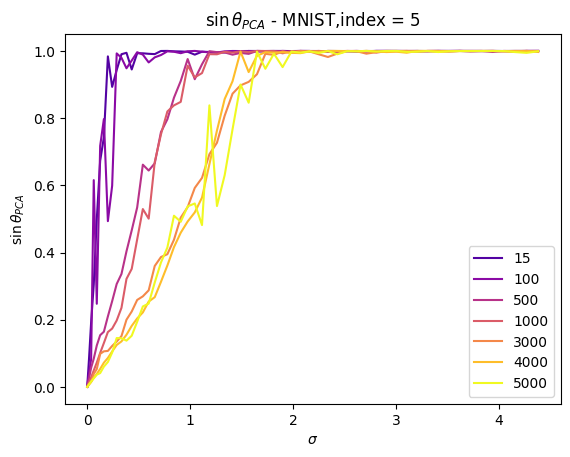}}
    \hfil
    \subfloat{\includegraphics[width=0.3\linewidth]{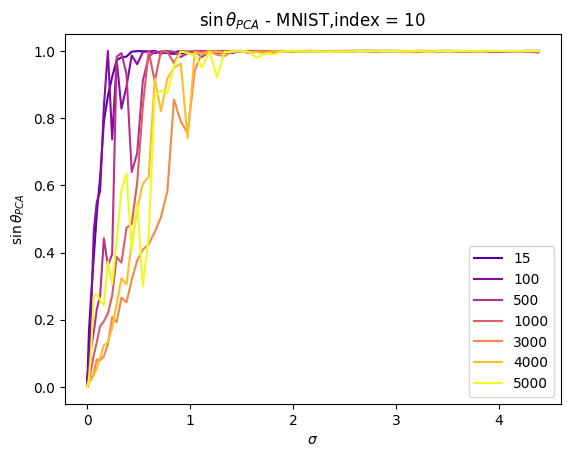}}
\caption{Effect of dataset size. The plots show $\sin{\theta_\text{PCA}}$ at different noise levels when trained on datasets with increasing size (lighter color). Each plot is of a different component index, for indices $0, 5, 10$ (left to right; index $0$ corresponds to the largest eigenvalue). Increasing the amount of training data improves the robustness to noise, and allows the appearance of high frequencies at higher noise levels, hence capturing more data nuances in the generated data and better generalization.} 
\label{fig:effect_of_dataset_size}
\end{figure}

In the linear case, \eqref{eq:sin_theta_pca_0} shows that the diffusion model's sample complexity is determined by the sample complexity of PCA, with a linear dependence on the dimension of the data. 
To further enhance our understanding of the relationship between the amount of training data and the generalization of the diffusion model, we repeat the experiment with varying datasets sizes. Figure \ref{fig:effect_of_dataset_size} shows the angle to noise profile for selected principal components, with the indices $0, 5, 10$ (left to right; index $0$ corresponds to the largest in a list of ordered eigenvalues). Increasing the amount of training data improves robustness to noise and enables the emergence of higher frequency components at higher noise levels, thereby capturing more nuances in the generated data.

\subsection{The Generated Distribution}\label{sec:output_distribution}
We now turn to discuss the distribution of the generated output, and how it relates to the natural data distribution. First, we analyze a generation process by repetitive denoising without additional noise, and show how it relates to power iteration. Then we discuss a similar process only with the injection of noise, reminiscent of other common sampling methods (e..g \citep{ddpm}). 
Given our linear model, the generation process is essentially the linear transformation given by the matrix
\begin{equation}\label{eq:linear_operator}
    \mathcal{P}_T = \Pi_{t=0}^T P_t
                  = P_0 \cdots P_t \cdots P_T.
\end{equation}
The generated output can be expressed as
\begin{equation}\label{eq:x_gen_no_injected_noise}
    x_g = \mathcal{P}_T \xi,
\end{equation}
where $\xi \sim \mathcal{N}(0, \sigma_T)$. 
Other than the visual aesthetic of the generated images, we are interested in their distribution, and how well it represents the natural distribution of train images. Thus, we would like to compare the generated covariance $\E{x_g x_g^\dag}$ to the natural covariance $\Sigma_0$.

In this context, a natural comparison is the power iteration (PI) method, which may be used to estimate the leading eigenvector of a matrix. This can be seen as another iterative form of generating data from random vectors. Unlike our projection, in PI we "project" a random vector onto the entire matrix, i.e. including the eigenvalues. In this case the denoiser would be $D^t_{PI} = \Sigma_0 \forall t$, where we ignore the normalization and focus on the direction of the final vector, since there is no normalization constraint for generated data in diffusion models.

We now turn to show how the reverse process performed by a repeated denoising as in \eqref{eq:linear_operator} converges in mean to PI. To this end, we make the following assumptions.

\begin{figure}
    \centering
    \subfloat{\includegraphics[width=0.3\linewidth]{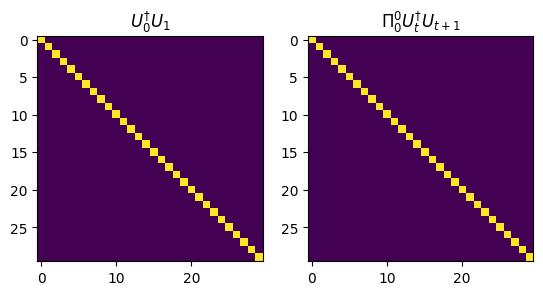}}
    \hfil
    \subfloat{\includegraphics[width=0.3\linewidth]{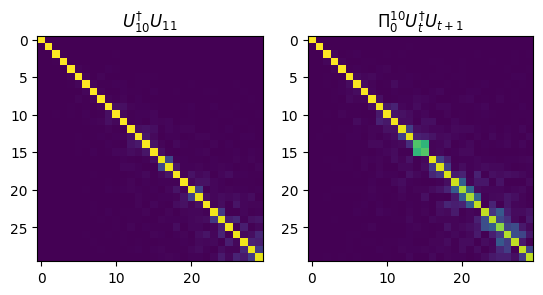}}
    \hfil
    \subfloat{\includegraphics[width=0.3\linewidth]{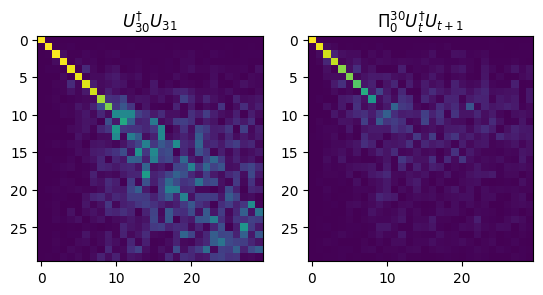}}
    \vfill
    \subfloat{\includegraphics[width=0.3\linewidth]{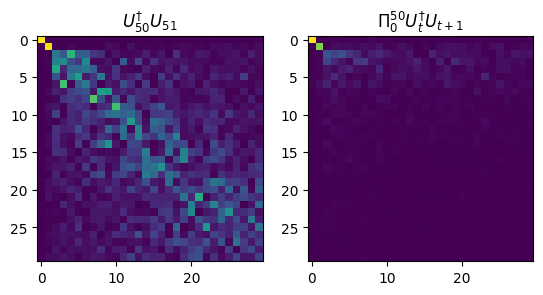}}
    \hfil
    \subfloat{\includegraphics[width=0.3\linewidth]{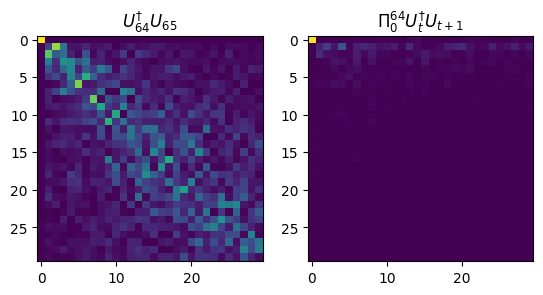}}
    \hfil
    \caption{The time point basis correlation matrices $U_\tau^T U_{\tau+1}$ (left per pair), together with the partial product $\Pi_{t=0}^\tau (U_t^\dag U_{t+1})$ (right per pair) at different time points. This justifies Assumption \ref{assumption:cross_products}, and shows that the total projection (bottom right image, for $\tau = T$) converges to the first eigenvector, similarly to the power method.}
    \label{fig:basis_correlation}
\end{figure}

\begin{assumption}\label{assumption:diagonal_elements}
Assume that \eqref{eq:sin_theta_pca_0} holds for all eigenvectors, i.e.,
\begin{equation}
    \E \sqrt{1 - \langle u_i^t, u_i \rangle^2} \approx \frac{\bar{\sigma}_t}{\lambda_i} \sqrt{\frac{d}{n}},
\end{equation}
for $i = , \dots, r - 1$.
\end{assumption}
This assumption is the extension of \eqref{eq:sin_theta_pca_0} to higher ranks, and is motivated by our simulations (Fig. \ref{fig:sin_theta_pca}). In addition, we make the following assumption regarding the cross products of components of different indices, at consecutive time steps. 
\begin{assumption}\label{assumption:cross_products}
For each index $i$ there exists a time $\tau_i$, where for $t \leq \tau_i$ and $j \leq i$,
\begin{equation}
    \E \langle u_i^t, u_j^{t + 1} \rangle = 0.
\end{equation}
In addition, $\tau_i > \tau_j$ for $i < j$.
\end{assumption}
This assumption is supported by our simulations in Fig. \ref{fig:basis_correlation}, and will be further discussed hereafter. Assumptions \ref{assumption:cross_products}, \ref{assumption:diagonal_elements} are an extension of \citet{nadler2008finite} to higher ranks. We leave their explicit derivation to future work, and focus on their implications to linear diffusion.

We are now ready to state our main result.
\begin{theorem}[Convergence to Power Iteration]
    Let $\sigma_t = \frac{1}{T}$, $t = 0, \dots, T$. Assuming \ref{assumption:cross_products}, \ref{assumption:diagonal_elements}, in the limit $T \to \infty$,
    \begin{equation}
        \E {x_g x_g^\dag} \propto u_0 u_0^\dag.
    \end{equation}
\end{theorem}
 
\begin{proof}
Let us analyze the product in \eqref{eq:linear_operator} to show how it relates to the power method. The linear operator representing the reverse process can be written as
\begin{equation}
    \mathcal{P}_T = U_0  \Pi_{t=0}^{T-1} (U_t^\dag U_{t+1})  U_T^\dag.
\end{equation}

The matrix product $U_t^\dagger U_{t+1}$ can be analyzed using the extension of \eqref{eq:sin_theta_pca_0} to higher ranks. Given \ref{assumption:diagonal_elements}, the expected inner product with the natural data component $u_i = u_i^{t=0}$ is given by
\begin{equation}
    \E \langle u_i^t, u_i \rangle \approx 1 - \frac{\bar{\sigma}^2_t}{\lambda_i^2} \frac{d}{n}. 
\end{equation}

\begin{figure}[t]
    \centering
    \begin{tikzpicture}
        \draw (2, 0) arc[start angle=0, end angle=60, radius=2];
    
        \draw[->, thick] (0,0) -- (2,0) node[right] {$\mathbf{u}$};
        \draw[->, thick] (0,0) -- (1, 1.732) node[above] {$\mathbf{u_{t+1}}$};
        \draw[->, thick] (0,0) -- (1.732, 1) node[above right] {$\mathbf{u_t}$};
    
        \draw[-, dashed] (1.732, 1) -- (1.732, 0);
        
        \draw[<->] (0.0, -0.1) -- (1.72, -0.1);
        \node at (0.8, -0.5) {$1 - \frac{\bar{\sigma}^2_t}{\lambda} \frac{d}{n}$};
    
        \draw[-] (0.5, 0) arc[start angle=0, end angle=30, radius=0.5] node at (0.7, 0.2) {$\theta_t$};
        \draw[-] (1, 0) arc[start angle=0, end angle=60, radius=1] node at (0.95, 0.86) {$\theta_{t + 1}$};
    \end{tikzpicture}
    \caption{Schematic illustration of the basis perturbation, per index.}
    \label{fig:theta_scheme}
\end{figure}
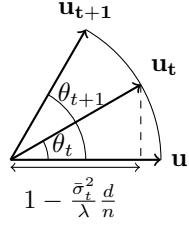
The evolution of this product over time is depicted in Figure \ref{fig:theta_scheme}. We are interested in the projection of $u_{t+1}$ onto $u_t$, which is the cosine of the angle $\Delta \theta = \theta_{t + 1} - \theta_t$. This angle is tractable for small noise levels, so we divide our analysis to two parts: $0 \leq t \leq \tau$ and $\tau \leq t \leq T$, where the choice of $\tau$ will soon be motivated.

First, we inspect the limit of $t \to 0$ ($0 \leq t \leq \tau$). For small angles, we can write
\begin{align}
    \Delta \theta =
    \arccos \Big{(} 1 - \frac{\bar{\sigma}^2_{t+1}}{\lambda^2} \frac{d}{n} \Big{)} -  \arccos \Big{(} 1 - \frac{\bar{\sigma}^2_t}{\lambda^2} \frac{d}{n} \Big{)}\approx
    \frac{d}{\lambda^2 n} (\bar{\sigma}^2_{t+1} - \bar{\sigma}^2_{t}) =
    \frac{\sigma_{t+1}^2 d}{\lambda^2 n},
\end{align}
since $\arccos{\theta} \approx \frac{\pi}{2} - \theta$ and $\bar{\sigma}_t^2 = \sum_{\tau=0}^t \sigma_\tau^2$. 
The diagonal elements in $U_t^\dagger U_{t+1}$ are then given by 
\begin{equation}
    \E \langle u_i^t, u^{t+1}_i \rangle \approx \cos{\frac{\sigma_{t+1}^2 d}{\lambda_i^2 n}},
\end{equation}
where the off-diagonal elements are negligible, since 
\begin{equation}\label{eq:u_cross_index}
    \E \langle u_i^t, u^{t+1}_j \rangle \approx
    \E \langle u_i^t, u^{t}_j \rangle = 0,
\end{equation}
which holds for $t \leq \tau_{r - 1}$ by Assumption \ref{assumption:cross_products}. Notice, that in small angles, $\langle u_i^t - u_i^{t + 1}, u \rangle = (\sigma_{t+1}^2 d) / (\lambda^2 n) \to 0$, so the vectors $u_i^t$ are co planar, as depicted in Figure \ref{fig:theta_scheme}.
Thus, the time point basis correlations $U_t^\dag U_{t+1}$ form an approximately diagonal matrix with the fraction $c_i \triangleq \cos{\frac{\sigma_{t+1}^2 d}{\lambda_i^2 n}}$ on the diagonal, where $c_i > c_j$ for $i < j$. We eliminate the dependence of $c_i$ on $t$ by choosing the constant schedule $\sigma_t = 1 / T \ \forall t$, to simplify the proof. However, many schedules can be used, as long as $c_{i, t} > c_{j, t}$ remains correct. 
Define the partial linear diffusion operator until time $\tau$ by $\E \mathcal{P}_\tau = \Pi_{t=0}^\tau P_t$. Then
\begin{align}\label{eq:convergence_to_PI}
    \E \mathcal{P}_\tau &= U_0 
                    \begin{pmatrix}
                        c_0^\tau & &  \\
                         &  \ddots & \\ 
                        && c_{r - 1}^\tau
                    \end{pmatrix} 
                    U_\tau^\dag 
                  = U_0 c_0^\tau
                    \begin{pmatrix}
                        1 & &  \\ 
                         & \big{(} \frac{c_1}{c_0}\big{)}^\tau & \\
                        && \ddots
                    \end{pmatrix}
                    U_\tau^\dag 
                \underset{\tau \to T}{\to} U_0 
                    \begin{pmatrix}
                        c_0^\tau & &  \\
                         &  0 \\ 
                        && \ddots
                    \end{pmatrix}
                    U_\tau^\dag,
\end{align}
where the diagonal elements decay as $\tau$ grows larger, as $c_i > c_j$ for $i < j$. Similarly to power iteration, the convergence rate depends on the ratio $c_1 / c_0$. The convergence rate might not be fast enough for the process to converge while the small angles approximation still holds. Thus, we continue with the second phase of our analysis, showing the convergence of the full reverse process.

We now turn to analyse the phase where $\tau \leq t \leq T$. In high noise levels, the correlation with the natural basis is low, and the products $U_t^\dag U_{t+1}$ are not exactly diagonal. However, the correlation "leaks" to a close neighborhood of the original component and the temporal products are still somewhat concentrated around their diagonal. This process happens in accordance with \eqref{assumption:diagonal_elements}, where the spreading of the diagonal elements happens for high indices in lower values of $t$ (less noise is needed to spread the correlation). This leads us to Assumption \ref{assumption:cross_products}, claiming that for each index $i$ there exists a time $\tau_i$ after which the small angle approximation does not hold; $\tau_i > \tau_j$ for $i < j$. This is apparent in practice, and depicted in \ref{fig:basis_correlation} (left image per duo). However, given the decaying diagonal structure of the partial operator $\mathcal{P}_\tau$, we will now show that \ref{assumption:cross_products} is sufficient for the total operator to converge as desired.

Suppose we added one more matrix multiplication to our former analysis, i.e. observe
\begin{equation}
    \E \mathcal{P}_\tau U_{\tau + 1} = 
                    U_0 c_0^\tau
                    \begin{pmatrix}
                        1 & &  \\ 
                         & \big{(} \frac{c_1}{c_0}\big{)}^\tau & \\
                        && \ddots
                    \end{pmatrix}
                    U_\tau^\dag U_{\tau + 1}.
\end{equation}
Assumption \ref{assumption:cross_products} guarantees $U_\tau^\dag U_{\tau+1}$ is diagonal just enough not to spoil the diagonality of the next partial operator $\E \mathcal{P}_{\tau + 1}$. Since the elements of the partial product $\E \mathcal{P}_\tau$ decay faster with $i$ than any single product $U_\tau^\dag U_{\tau+1}$, $\E \mathcal{P}_{\tau + 1}$ is also diagonal. Overall, the final product is a diagonal matrix with a spectrum that converges to be concentrated around the first eigenvalue, where we can control the distribution of the generated data by the choice of the diffusion parameters. Figure \ref{fig:basis_correlation} shows our simulation of the process, supporting both assumption \ref{assumption:cross_products} and the result stated by this theorem.
\end{proof}

Oftentimes, the reverse process includes the injection of noise to the intermediate images (e.g. \citep{ddpm}). The overall transformation in this case is given by
\begin{equation}\label{eq:generation_with_injected_noise}
     x_g = \Sigma_{t=0}^T \Pi_{\tau=0}^t P_\tau \xi_t
         = P_0 \cdots P_T \xi_T + \dots + P_0 \xi_0
\end{equation}
for some schedule $\{ \xi_t \}_{t=0}^T$ (for example, $\xi_T \sim \mathcal{N}(0, 1)$ and $\xi_t = \mathcal{N}(0, 1/T)$ for $t = 0, \dots, T - 1$). 
In this case, the generated output is a combination of a (purely) noisy image that was repeatedly correlated to converge to $v_0$ (as shown above), with generally lower noise levels that are "lightly" correlated, although to the cleaner projection operators. The generated output can thus be seen as a combination of three conceptual parts, with a different balance of the noise level and the portrayed components. 

\paragraph{The first eigenvector}
The first part of the sum in \eqref{eq:generation_with_injected_noise} is $P_0 \cdots P_T \xi_T$,  the estimation of the eigenvector with the largest eigenvalue, as shown above theoretically in \eqref{eq:convergence_to_PI} and empirically in the right matrix of the bottom right duo in Fig. \ref{fig:basis_correlation}. The "strongest" noise is repeatedly correlated to be concentrated around the first eigenvector.

\paragraph{The entire (clean) spectrum}
The last part in \eqref{eq:generation_with_injected_noise} is $P_0 \xi_0$, a weak noise level that is spread across all components. This noise is very lightly and not repeatedly correlated, although to a clean version of the natural data basis. 

\paragraph{In between}
The third part consists of all the intermediate products $\Pi_{\tau=0}^t P_\tau \xi_t$. The product operators $\Pi_{\tau=0}^t P_\tau$ preserve varying parts of the natural spectrum, according to $t$ - as $t$ grows, the total projection tends to retain only the components associated with larger eigenvalues. This can bee seen in Fig. \ref{fig:basis_correlation}. The right matrix in each pair shows the product $\Pi_{\tau=0}^t P_\tau$ for varying values of $t$. The total projections range from the entire spectrum (top left), to only the leading eigenvalue (bottom right). In between, the products are diagonal matrices where the entries in the indices of the smaller eigenvalues have already diminished, in a similar way to the convergence described in \eqref{eq:convergence_to_PI}.

Thus, we get a combination of a solid estimation of the leading eigenvector, together with a more uniform and week sampling of the components with low eigenvalues in the natural data basis. In between, the intermediate projections are at different levels of convergence to the leading eigenvector, hence tend to be more concentrated on components with large eigenvalues as $t \to T$. The freedom in choice of schedule $\{ \xi_t \}_{t=0}^T$, allows control of the spread of the final distribution on the natural data components.

\begin{figure}[!h]
\centering
    \subfloat{\includegraphics[width=0.3\linewidth]{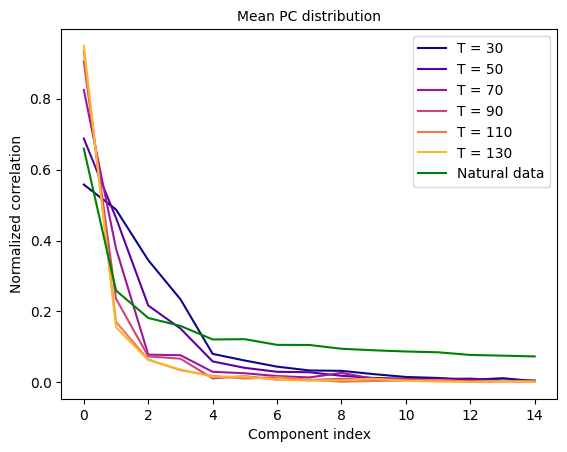}}
    \hfil
    \subfloat{\includegraphics[width=0.3\linewidth]{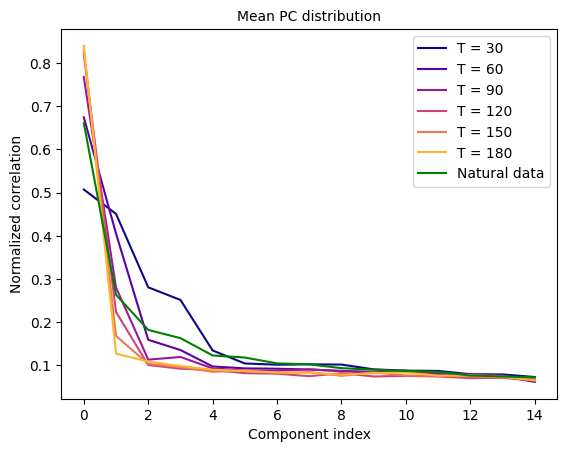}}
    \hfil
    \subfloat{\includegraphics[width=0.3\linewidth]{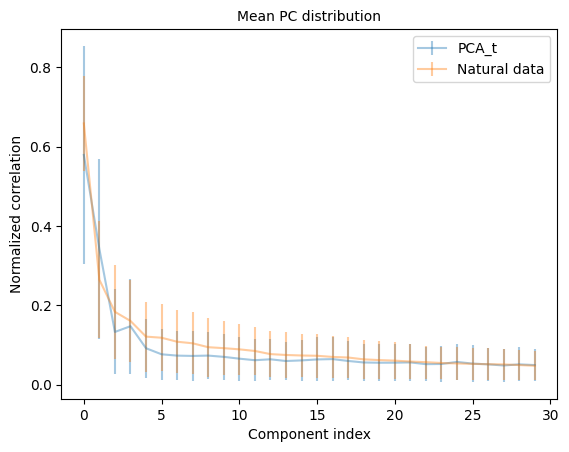}}
\caption{The empirical distribution of generated images over the natural principal components, with (middle) and without (left) injected noise. On the right - the best configuration with the generated standard deviation (see Sec. \ref{sec:output_distribution}).} 
\label{fig:mean_PC_distribution}
\end{figure}

To inspect this, we can plot the empirical absolute distribution of generated images over the clean PCs, given by
\begin{equation}\label{eq:c_i}
    c_i = \frac{1}{n} \sum_{j = 1}^n \frac{|\langle u_i,  x_j\rangle|}{\norm{x_j}_2},
\end{equation}
where $v_i$ is the clean principal component with index $i$ (defined in \eqref{eq:spiked_model}). 
Figure \ref{fig:mean_PC_distribution} shows the empirical distribution of generated images over the clean principal components. On the left, we plot the distribution without injected noise, i.e. $x_g = \mathcal{P}_T \xi$ (as in \eqref{eq:x_gen_no_injected_noise}), for various values of $T$. As we show above, the distribution tends to be concentrated on the first eigenvector as $T$ increases. The center plot shows the distribution of the process including the injected noise in the intermediate denoising steps. While in the low indices the dominant behavior is similar to the former case, the higher orders do not converge to zero and maintain their presence in the generated distribution. On the right, we picked the best configuration ($T = 65$ in this case) to approximate the natural distribution. In addition to the convergence in mean, we included the standard deviation of the natural and generated samples, resulting in a decent fit to the target distribution.

\section{Empirical Extension to Deep Denoisers}\label{sec:zero_bias_denoisers}
In the linear case described above, the optimal denoiser is given by the PCA projection onto the clean(er) data. These denoisers are computed with the training data, and their principal components do not depend on the input in the reverse process. When the denoiser is nonlinear, and might be implemented using a deep neural network, its input-output mapping can be locally expressed via the network Jacobian, by
\begin{equation}\label{eq:Jacobian}
    D(x_t) = \nabla D(x_t) x_t = V_t \Lambda_t V_t^\dag x_t,
\end{equation}
where $V_t \Lambda_t V_t^\dag$ denotes the eigen decomposition of the Jacobian calculated at $x_t$. For simplicity, we assume that the Jacobian is symmetric and non-negative (which is approximately true \citep{mohan2019robust}). Note that in this case, the denoising base depends on the input image (and noise level). While the network is non linear, we can follow the generation path in the sampling process and inspect the basis of the network Jacobians calculated at the intermediate sampled points $x_t$. We can then trace $\sin{\theta_J} = \sqrt{1 - \langle v_i^t, v^{t = 0}_i \rangle^2}$ where $v_i^t$ is the $i^{th}$ column in $V_t$ defined in \eqref{eq:Jacobian}, in a similar way to our simulations of the linear case (Figure \ref{fig:sin_theta_pca}). This can be calculated per generation path, where $x_0$ is the final generated image, and $V_0$ is the basis of the Jacobian calculated at this final point.

\begin{figure}[!h]
    \centering
    \subfloat{\includegraphics[width=0.45\linewidth]{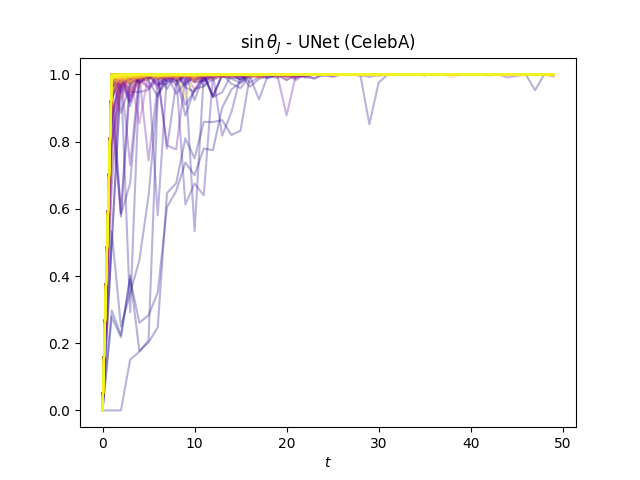}}
    \hfil
    \subfloat{\includegraphics[width=0.45\linewidth]{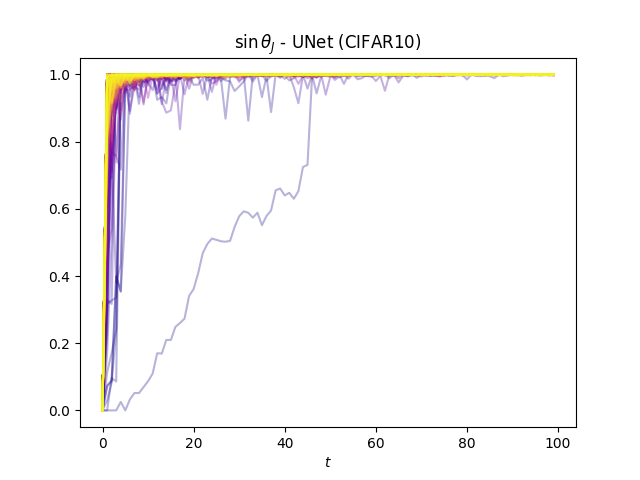}}
    \caption{Image generation - the sine of the angle between Jacobian eigenvectors at the final generated image ($t = 0$) and intermediate iterations ($t > 0$). The diffusion model includes a UNet-based denoiser trained on CelebA (left) or CIFAR10 (right). Color by index (the darker the color the lower the index, referring to columns of the Jacobian basis $V_t$). The Jacobians of the nonlinear denoiser conform to the behavior of the linear model.}
    \label{fig:sin_theta_unet_generation}
\end{figure}

Figure \ref{fig:sin_theta_unet_generation} shows $\sin{\theta_J}$ calculated using the Jacobians of a UNet based diffusion model, described in \citep{ning2023input}. This model was simply chosen as the \footnote{\url{https://paperswithcode.com/sota/image-generation-on-celeba-64x64}}{state-of-the-art} in the task of image generation considering the CelebA dataset at the time of writing this paper. We used the default settings and calculated the Jacobians at the final iterations. We plot the results for the leading 300 Jacobian eigenvectors, where the color is assigned by the index - darker colors for lower indices $i$. We repeated the experiment sampling images from the CelebA dataset (left) and CIFAR 10 (right). Even though the denoising model is far from linear, the decay of the angle between the denoising basis in high noise levels and the natural denoising basis is similar to the decay in the linear case (compare to Figure~\ref{fig:sin_theta_pca}). In this case as well, the correlation of the low indices (and hence low frequencies) withstands higher noise levels, thus appearing first in the generation process. As this is the basis of our analysis comparing the reverse diffusion process to power iteration, this experiment shows that our analysis is relevant in a broader context and not just in the simplified linear case.

\section{Conclusion}
In this paper, we discuss a simple diffusion model with a linear denoiser and normalization free sampler, that allows us to cast the diffusion problem as noisy PCA, and make the connection to the spiked covariance model assuming that the natural data distribution reside in a low dimensional subspace. This enables us to show that in the linear case, the generation process acts as a ``correlation machine'', where initial random noise is repeatedly correlated to noisy estimations of the natural data basis, to finally embody the true distribution, in a manner similar to the power iteration method. We show that in this process, low frequencies emerge earlier, and more data contributes to a richer representation per the same diffusion configuration. Finally, we demonstrate the relevance of our analysis also in a deep, non-linear diffusion denoiser.  

We acknowledge the limitation of admitting a linear model, with its lack of ability to represent the complex data often expected of diffusion models. While our theoretical setting is modest, we empirically demonstrate how our observations deduced from a simple linear model and classic theory \citep{johnstone2001distribution, nadler2008finite} are relevant to more general models and datasets. This enables us to shed light on the internal mechanism powering this technology, and connect it to a rich pool of theory and prevalent methods such as power iteration.   

\section*{Acknowledgments}
The authors extend their sincere gratitude to Mojtaba Ardakani, Tom Duerig and Shady Abu-Hussein for their valuable feedback. RG and DW were partially supported by the Israeli Innovation Authority, the Israeli council of higher education and the center for AI and Data Science at Tel University. RG and DW are also thankful for KLA and Google for their support.

\bibliography{bibtex} 

\begin{thebibliography}{47}
\providecommand{\natexlab}[1]{#1}
\providecommand{\url}[1]{\texttt{#1}}
\expandafter\ifx\csname urlstyle\endcsname\relax
  \providecommand{\doi}[1]{doi: #1}\else
  \providecommand{\doi}{doi: \begingroup \urlstyle{rm}\Url}\fi

\bibitem[Alcaraz \& Strodthoff(2023)Alcaraz and Strodthoff]{alcaraz2022diffusion}
Juan~Lopez Alcaraz and Nils Strodthoff.
\newblock Diffusion-based time series imputation and forecasting with structured state space models.
\newblock \emph{Transactions on Machine Learning Research}, 2023.
\newblock ISSN 2835-8856.
\newblock URL \url{https://openreview.net/forum?id=hHiIbk7ApW}.

\bibitem[Amit et~al.(2021)Amit, Shaharbany, Nachmani, and Wolf]{amit2021segdiff}
Tomer Amit, Tal Shaharbany, Eliya Nachmani, and Lior Wolf.
\newblock Segdiff: Image segmentation with diffusion probabilistic models.
\newblock \emph{arXiv preprint arXiv:2112.00390}, 2021.

\bibitem[Austin et~al.(2021)Austin, Johnson, Ho, Tarlow, and Van Den~Berg]{austin2021structured}
Jacob Austin, Daniel~D Johnson, Jonathan Ho, Daniel Tarlow, and Rianne Van Den~Berg.
\newblock Structured denoising diffusion models in discrete state-spaces.
\newblock \emph{Advances in Neural Information Processing Systems}, 34:\penalty0 17981--17993, 2021.

\bibitem[Baranchuk et~al.(2022)Baranchuk, Voynov, Rubachev, Khrulkov, and Babenko]{baranchuk2021label}
Dmitry Baranchuk, Andrey Voynov, Ivan Rubachev, Valentin Khrulkov, and Artem Babenko.
\newblock Label-efficient semantic segmentation with diffusion models.
\newblock In \emph{International Conference on Learning Representations}, 2022.
\newblock URL \url{https://openreview.net/forum?id=SlxSY2UZQT}.

\bibitem[Benton et~al.(2024)Benton, Bortoli, Doucet, and Deligiannidis]{iteration_complexity_benton2023linear}
Joe Benton, Valentin~De Bortoli, Arnaud Doucet, and George Deligiannidis.
\newblock Nearly \$d\$-linear convergence bounds for diffusion models via stochastic localization.
\newblock In \emph{The Twelfth International Conference on Learning Representations}, 2024.
\newblock URL \url{https://openreview.net/forum?id=r5njV3BsuD}.

\bibitem[Biroli \& M{\'e}zard(2023)Biroli and M{\'e}zard]{biroli2023physics}
Giulio Biroli and Marc M{\'e}zard.
\newblock Generative diffusion in very large dimensions.
\newblock \emph{Journal of Statistical Mechanics: Theory and Experiment}, 2023\penalty0 (9):\penalty0 093402, 2023.

\bibitem[Block et~al.(2020)Block, Mroueh, and Rakhlin]{block2020generative}
Adam Block, Youssef Mroueh, and Alexander Rakhlin.
\newblock Generative modeling with denoising auto-encoders and langevin sampling.
\newblock \emph{arXiv preprint arXiv:2002.00107}, 2020.

\bibitem[Bortoli(2022)]{low_dim_debortoli2022convergence}
Valentin~De Bortoli.
\newblock Convergence of denoising diffusion models under the manifold hypothesis.
\newblock \emph{Transactions on Machine Learning Research}, 2022.
\newblock ISSN 2835-8856.
\newblock URL \url{https://openreview.net/forum?id=MhK5aXo3gB}.
\newblock Expert Certification.

\bibitem[Brempong et~al.(2022)Brempong, Kornblith, Chen, Parmar, Minderer, and Norouzi]{brempong2022denoising}
Emmanuel~Asiedu Brempong, Simon Kornblith, Ting Chen, Niki Parmar, Matthias Minderer, and Mohammad Norouzi.
\newblock Denoising pretraining for semantic segmentation.
\newblock In \emph{Proceedings of the IEEE/CVF conference on computer vision and pattern recognition}, pp.\  4175--4186, 2022.

\bibitem[Cai et~al.(2020)Cai, Yang, Averbuch-Elor, Hao, Belongie, Snavely, and Hariharan]{cai2020learning}
Ruojin Cai, Guandao Yang, Hadar Averbuch-Elor, Zekun Hao, Serge Belongie, Noah Snavely, and Bharath Hariharan.
\newblock Learning gradient fields for shape generation.
\newblock In \emph{Computer Vision--ECCV 2020: 16th European Conference, Glasgow, UK, August 23--28, 2020, Proceedings, Part III 16}, pp.\  364--381. Springer, 2020.

\bibitem[Chen et~al.(2024)Chen, Zhou, Wang, Shen, and Lyu]{chen2023geometric}
Defang Chen, Zhenyu Zhou, Can Wang, Chunhua Shen, and Siwei Lyu.
\newblock On the trajectory regularity of {ODE}-based diffusion sampling.
\newblock In Ruslan Salakhutdinov, Zico Kolter, Katherine Heller, Adrian Weller, Nuria Oliver, Jonathan Scarlett, and Felix Berkenkamp (eds.), \emph{Proceedings of the 41st International Conference on Machine Learning}, volume 235 of \emph{Proceedings of Machine Learning Research}, pp.\  7905--7934. PMLR, 21--27 Jul 2024.
\newblock URL \url{https://proceedings.mlr.press/v235/chen24bm.html}.

\bibitem[Chen et~al.(2023{\natexlab{a}})Chen, Huang, Zhao, and Wang]{chen2023score}
Minshuo Chen, Kaixuan Huang, Tuo Zhao, and Mengdi Wang.
\newblock Score approximation, estimation and distribution recovery of diffusion models on low-dimensional data.
\newblock In \emph{International Conference on Machine Learning}, pp.\  4672--4712. PMLR, 2023{\natexlab{a}}.

\bibitem[Chen et~al.(2021)Chen, Zhang, Zen, Weiss, Norouzi, and Chan]{chen2020wavegrad}
Nanxin Chen, Yu~Zhang, Heiga Zen, Ron~J Weiss, Mohammad Norouzi, and William Chan.
\newblock Wavegrad: Estimating gradients for waveform generation.
\newblock In \emph{International Conference on Learning Representations}, 2021.
\newblock URL \url{https://openreview.net/forum?id=NsMLjcFaO8O}.

\bibitem[Chen et~al.(2023{\natexlab{b}})Chen, Daras, and Dimakis]{iteration_complexity_chen2023restoration}
Sitan Chen, Giannis Daras, and Alex Dimakis.
\newblock Restoration-degradation beyond linear diffusions: A non-asymptotic analysis for ddim-type samplers.
\newblock In \emph{International Conference on Machine Learning}, pp.\  4462--4484. PMLR, 2023{\natexlab{b}}.

\bibitem[Delbracio \& Milanfar(2023)Delbracio and Milanfar]{indi}
Mauricio Delbracio and Peyman Milanfar.
\newblock Inversion by direct iteration: An alternative to denoising diffusion for image restoration.
\newblock \emph{Transactions on Machine Learning Research}, 2023.
\newblock ISSN 2835-8856.
\newblock URL \url{https://openreview.net/forum?id=VmyFF5lL3F}.
\newblock Featured Certification.

\bibitem[Devroye et~al.(2018)Devroye, Mehrabian, and Reddad]{devroye2018total}
Luc Devroye, Abbas Mehrabian, and Tommy Reddad.
\newblock The total variation distance between high-dimensional gaussians with the same mean.
\newblock \emph{arXiv preprint arXiv:1810.08693}, 2018.

\bibitem[Dhariwal \& Nichol(2021)Dhariwal and Nichol]{dhariwal2021diffusion}
Prafulla Dhariwal and Alexander Nichol.
\newblock Diffusion models beat gans on image synthesis.
\newblock \emph{Advances in neural information processing systems}, 34:\penalty0 8780--8794, 2021.

\bibitem[Ho et~al.(2020)Ho, Jain, and Abbeel]{ddpm}
Jonathan Ho, Ajay Jain, and Pieter Abbeel.
\newblock Denoising diffusion probabilistic models.
\newblock \emph{Advances in neural information processing systems}, 33:\penalty0 6840--6851, 2020.

\bibitem[Hoogeboom et~al.(2021)Hoogeboom, Nielsen, Jaini, Forr{\'e}, and Welling]{hoogeboom2021argmax}
Emiel Hoogeboom, Didrik Nielsen, Priyank Jaini, Patrick Forr{\'e}, and Max Welling.
\newblock Argmax flows and multinomial diffusion: Learning categorical distributions.
\newblock \emph{Advances in Neural Information Processing Systems}, 34:\penalty0 12454--12465, 2021.

\bibitem[Huang et~al.(2024)Huang, Huang, and Lin]{iteration_complexity_huang2024convergence}
Daniel~Zhengyu Huang, Jiaoyang Huang, and Zhengjiang Lin.
\newblock Convergence analysis of probability flow ode for score-based generative models.
\newblock \emph{arXiv preprint arXiv:2404.09730}, 2024.

\bibitem[Johnstone(2001)]{johnstone2001distribution}
Iain~M Johnstone.
\newblock On the distribution of the largest eigenvalue in principal components analysis.
\newblock \emph{The Annals of statistics}, 29\penalty0 (2):\penalty0 295--327, 2001.

\bibitem[Kadkhodaie et~al.(2024)Kadkhodaie, Guth, Simoncelli, and Mallat]{kadkhodaie2023generalization}
Zahra Kadkhodaie, Florentin Guth, Eero~P Simoncelli, and St{\'e}phane Mallat.
\newblock Generalization in diffusion models arises from geometry-adaptive harmonic representations.
\newblock In \emph{The Twelfth International Conference on Learning Representations}, 2024.
\newblock URL \url{https://openreview.net/forum?id=ANvmVS2Yr0}.

\bibitem[Karras et~al.(2022)Karras, Aittala, Aila, and Laine]{karras2022elucidating}
Tero Karras, Miika Aittala, Timo Aila, and Samuli Laine.
\newblock Elucidating the design space of diffusion-based generative models.
\newblock \emph{Advances in neural information processing systems}, 35:\penalty0 26565--26577, 2022.

\bibitem[Kong et~al.(2021)Kong, Ping, Huang, Zhao, and Catanzaro]{kong2020diffwave}
Zhifeng Kong, Wei Ping, Jiaji Huang, Kexin Zhao, and Bryan Catanzaro.
\newblock Diffwave: A versatile diffusion model for audio synthesis.
\newblock In \emph{International Conference on Learning Representations}, 2021.
\newblock URL \url{https://openreview.net/forum?id=a-xFK8Ymz5J}.

\bibitem[Li \& Yan(2024)Li and Yan]{low_dim_li2024adapting}
Gen Li and Yuling Yan.
\newblock Adapting to unknown low-dimensional structures in score-based diffusion models.
\newblock \emph{arXiv preprint arXiv:2405.14861}, 2024.

\bibitem[Li et~al.(2024{\natexlab{a}})Li, Wei, Chen, and Chi]{iteration_complexity_li2023towards}
Gen Li, Yuting Wei, Yuxin Chen, and Yuejie Chi.
\newblock Towards non-asymptotic convergence for diffusion-based generative models.
\newblock In \emph{The Twelfth International Conference on Learning Representations}, 2024{\natexlab{a}}.
\newblock URL \url{https://openreview.net/forum?id=4VGEeER6W9}.

\bibitem[Li et~al.(2024{\natexlab{b}})Li, Wei, Chi, and Chen]{iteration_complexity_li2024sharp}
Gen Li, Yuting Wei, Yuejie Chi, and Yuxin Chen.
\newblock A sharp convergence theory for the probability flow odes of diffusion models.
\newblock \emph{arXiv preprint arXiv:2408.02320}, 2024{\natexlab{b}}.

\bibitem[Li et~al.(2022)Li, Thickstun, Gulrajani, Liang, and Hashimoto]{li2022diffusion}
Xiang Li, John Thickstun, Ishaan Gulrajani, Percy~S Liang, and Tatsunori~B Hashimoto.
\newblock Diffusion-lm improves controllable text generation.
\newblock \emph{Advances in Neural Information Processing Systems}, 35:\penalty0 4328--4343, 2022.

\bibitem[Milanfar(2013)]{milanfar2013atour}
Peyman Milanfar.
\newblock A tour of modern image filtering: New insights and methods, both practical and theoretical.
\newblock \emph{IEEE Signal Processing Magazine}, 30\penalty0 (1):\penalty0 106--128, 2013.
\newblock \doi{10.1109/MSP.2011.2179329}.

\bibitem[Milanfar \& Delbracio(2024)Milanfar and Delbracio]{milanfar2024denoising}
Peyman Milanfar and Mauricio Delbracio.
\newblock Denoising: A powerful building-block for imaging, inverse problems, and machine learning.
\newblock \emph{arXiv preprint arXiv:2409.06219}, 2024.

\bibitem[Mohan et~al.(2020)Mohan, Kadkhodaie, Simoncelli, and Fernandez-Granda]{mohan2019robust}
Sreyas Mohan, Zahra Kadkhodaie, Eero~P. Simoncelli, and Carlos Fernandez-Granda.
\newblock Robust and interpretable blind image denoising via bias-free convolutional neural networks.
\newblock In \emph{International Conference on Learning Representations}, 2020.
\newblock URL \url{https://openreview.net/forum?id=HJlSmC4FPS}.

\bibitem[Nadler(2008)]{nadler2008finite}
Boaz Nadler.
\newblock {Finite sample approximation results for principal component analysis: A matrix perturbation approach}.
\newblock \emph{The Annals of Statistics}, 36\penalty0 (6):\penalty0 2791 -- 2817, 2008.

\bibitem[Ning et~al.(2023)Ning, Sangineto, Porrello, Calderara, and Cucchiara]{ning2023input}
Mang Ning, Enver Sangineto, Angelo Porrello, Simone Calderara, and Rita Cucchiara.
\newblock Input perturbation reduces exposure bias in diffusion models.
\newblock In \emph{Proceedings of the 40th International Conference on Machine Learning}, ICML'23. JMLR.org, 2023.

\bibitem[Oko et~al.(2023)Oko, Akiyama, and Suzuki]{low_dim_oko2023diffusion}
Kazusato Oko, Shunta Akiyama, and Taiji Suzuki.
\newblock Diffusion models are minimax optimal distribution estimators.
\newblock In \emph{International Conference on Machine Learning}, pp.\  26517--26582. PMLR, 2023.

\bibitem[Rasul et~al.(2021)Rasul, Sheikh, Schuster, Bergmann, and Vollgraf]{rasul2021multivariate}
Kashif Rasul, Abdul-Saboor Sheikh, Ingmar Schuster, Urs~M Bergmann, and Roland Vollgraf.
\newblock Multivariate probabilistic time series forecasting via conditioned normalizing flows.
\newblock In \emph{International Conference on Learning Representations}, 2021.
\newblock URL \url{https://openreview.net/forum?id=WiGQBFuVRv}.

\bibitem[Savinov et~al.(2022)Savinov, Chung, Binkowski, Elsen, and van~den Oord]{savinov2021step}
Nikolay Savinov, Junyoung Chung, Mikolaj Binkowski, Erich Elsen, and Aaron van~den Oord.
\newblock Step-unrolled denoising autoencoders for text generation.
\newblock In \emph{International Conference on Learning Representations}, 2022.
\newblock URL \url{https://openreview.net/forum?id=T0GpzBQ1Fg6}.

\bibitem[Sohl-Dickstein et~al.(2015)Sohl-Dickstein, Weiss, Maheswaranathan, and Ganguli]{sohl2015deep}
Jascha Sohl-Dickstein, Eric Weiss, Niru Maheswaranathan, and Surya Ganguli.
\newblock Deep unsupervised learning using nonequilibrium thermodynamics.
\newblock In \emph{International conference on machine learning}, pp.\  2256--2265. PMLR, 2015.

\bibitem[Song et~al.(2021{\natexlab{a}})Song, Meng, and Ermon]{ddim}
Jiaming Song, Chenlin Meng, and Stefano Ermon.
\newblock Denoising diffusion implicit models.
\newblock In \emph{International Conference on Learning Representations}, 2021{\natexlab{a}}.
\newblock URL \url{https://openreview.net/forum?id=St1giarCHLP}.

\bibitem[Song \& Ermon(2019)Song and Ermon]{song2019generative}
Yang Song and Stefano Ermon.
\newblock Generative modeling by estimating gradients of the data distribution.
\newblock \emph{Advances in neural information processing systems}, 32, 2019.

\bibitem[Song \& Ermon(2020)Song and Ermon]{song2020improved}
Yang Song and Stefano Ermon.
\newblock Improved techniques for training score-based generative models.
\newblock \emph{Advances in neural information processing systems}, 33:\penalty0 12438--12448, 2020.

\bibitem[Song et~al.(2021{\natexlab{b}})Song, Durkan, Murray, and Ermon]{song2021maximum}
Yang Song, Conor Durkan, Iain Murray, and Stefano Ermon.
\newblock Maximum likelihood training of score-based diffusion models.
\newblock \emph{Advances in neural information processing systems}, 34:\penalty0 1415--1428, 2021{\natexlab{b}}.

\bibitem[Song et~al.(2021{\natexlab{c}})Song, Sohl-Dickstein, Kingma, Kumar, Ermon, and Poole]{song2020score}
Yang Song, Jascha Sohl-Dickstein, Diederik~P Kingma, Abhishek Kumar, Stefano Ermon, and Ben Poole.
\newblock Score-based generative modeling through stochastic differential equations.
\newblock In \emph{International Conference on Learning Representations}, 2021{\natexlab{c}}.
\newblock URL \url{https://openreview.net/forum?id=PxTIG12RRHS}.

\bibitem[Talebi \& Milanfar(2014)Talebi and Milanfar]{talebi2014nonlocal}
Hossein Talebi and Peyman Milanfar.
\newblock Nonlocal image editing.
\newblock \emph{IEEE Transactions on Image Processing}, 23\penalty0 (10):\penalty0 4460--4473, 2014.
\newblock \doi{10.1109/TIP.2014.2348870}.

\bibitem[Talebi \& Milanfar(2016)Talebi and Milanfar]{talebi2016fast}
Hossein Talebi and Peyman Milanfar.
\newblock Fast multilayer laplacian enhancement.
\newblock \emph{IEEE Transactions on Computational Imaging}, 2\penalty0 (4):\penalty0 496--509, 2016.
\newblock \doi{10.1109/TCI.2016.2607142}.

\bibitem[Tashiro et~al.(2021)Tashiro, Song, Song, and Ermon]{tashiro2021csdi}
Yusuke Tashiro, Jiaming Song, Yang Song, and Stefano Ermon.
\newblock Csdi: Conditional score-based diffusion models for probabilistic time series imputation.
\newblock \emph{Advances in Neural Information Processing Systems}, 34:\penalty0 24804--24816, 2021.

\bibitem[Wang et~al.(2024)Wang, Zhang, Zhang, Chen, Ma, and Qu]{wang2024diffusion}
Peng Wang, Huijie Zhang, Zekai Zhang, Siyi Chen, Yi~Ma, and Qing Qu.
\newblock Diffusion models learn low-dimensional distributions via subspace clustering.
\newblock \emph{arXiv preprint arXiv:2409.02426}, 2024.

\bibitem[Yu et~al.(2022)Yu, Xie, Ma, Jia, Pang, Gao, Zhu, Zhu, and Wu]{yu2022latent}
Peiyu Yu, Sirui Xie, Xiaojian Ma, Baoxiong Jia, Bo~Pang, Ruiqi Gao, Yixin Zhu, Song-Chun Zhu, and Ying~Nian Wu.
\newblock Latent diffusion energy-based model for interpretable text modeling.
\newblock In \emph{Proceedings of International Conference on Machine Learning (ICML)}, July 2022.

\end{thebibliography}
\bibliographystyle{iclr2025_conference}

\end{document}